\documentclass{article}
\usepackage{ijcai22}

\usepackage{times}

\usepackage{soul}
\usepackage{url}
\usepackage[hidelinks]{hyperref}
\usepackage[utf8]{inputenc}
\usepackage[small]{caption}
\usepackage{graphicx}
\usepackage{amsmath}
\usepackage{booktabs}
\usepackage{amssymb}
\usepackage{algorithm}
\usepackage{algorithmic}
\newcommand{\tabelarraystretch}{1}

\usepackage{mathrsfs}
\usepackage{color}

\title{Contrastive Multi-view Hyperbolic Hierarchical Clustering}

\author{
Fangfei Lin$^{1,2}$\footnote{This work was
done when Fangfei Lin was an intern at Tencent.}\and Bing Bai$^2$\and Kun Bai$^2$\and Yazhou Ren$^1$\and Peng Zhao$^1$\And Zenglin Xu$^{3,4}$\footnote{Corresponding Author}\\
\affiliations
$^1$University of Electronic Science and Technology of China, Chengdu, China\\
$^2$Tencent Security Big Data Lab, Tencent Inc., China\\
$^3$Harbin Institute of Technology, Shenzhen, China\\
$^4$Department of Network Intelligence,
Peng Cheng National Lab, Shenzhen, China\\
\emails
phoebe.lin1108@gmail.com,
\{icebai, kunbai\}@tencent.com,
yazhou.ren@uestc.edu.cn,
zhaop211@gmail.com,
zenglin@gmail.com
}

\begin{document}

\maketitle

\begin{abstract}
Hierarchical clustering recursively partitions data at an increasingly finer granularity.
In real-world applications, multi-view data have become increasingly important.
This raises a less investigated problem, i.e., multi-view hierarchical clustering, to better understand the hierarchical structure of multi-view data.
To this end, we propose a novel neural network-based model, namely Contrastive Multi-view Hyperbolic Hierarchical Clustering~(CMHHC).~It consists of three components, i.e., multi-view alignment learning, aligned feature similarity learning, and continuous hyperbolic hierarchical clustering.
First, we align sample-level representations across multiple views in a contrastive way to capture the view-invariance information.
Next, we utilize both the manifold and Euclidean similarities to improve the metric property.
Then, we embed the representations into a hyperbolic space and optimize the hyperbolic embeddings via a continuous relaxation of hierarchical clustering loss.
Finally, a binary clustering tree is decoded from optimized hyperbolic embeddings.
Experimental results on five real-world datasets demonstrate the effectiveness of the proposed method and its components.
\end{abstract}

\section{Introduction}
Clustering is one of the fundamental problems in data analysis, which aims to categorize unlabeled data points into clusters. 
Existing clustering methods can be divided into partitional clustering and hierarchical clustering~(HC)~\cite{jain1999data}. The difference is that partitional clustering produces only one partition, while hierarchical clustering produces a nested series of partitions.
Compared with partitional clustering, hierarchical clustering reveals more information about fine-grained similarity relationships and structures of data.

Hierarchical clustering has gained intensive attention. There are usually two types of hierarchical methods, i.e., the agglomerative methods and continuous ones. Agglomerative methods include Single-Linkage, Complete-Linkage, Average-Linkage, Ward-Linkage, etc.; continuous methods include Ultrametric Fitting (UFit)~\cite{chierchia2020ultrametric} and Hyperbolic Hierarchical Clustering~(HypHC)~\cite{chami2020trees}, etc.
Existing HC methods are usually limited to data from a single source. However, with the advances of data acquisition in real-world applications, data of an underlying signal is often collected from heterogeneous sources or feature subsets~\cite{li2018survey}.
For example, images can be described by the local binary pattern~(LBP) descriptor and the scale-invariant feature transform~(SIFT) descriptor; websites can be represented by text, pictures, and other structured metadata.
Data from different views may contain complementary and consensus information. Hence it would be greatly beneficial to utilize multi-view data to fertilize hierarchical clustering.

\begin{figure}[ht]
	\centering
	\includegraphics[width=\linewidth]{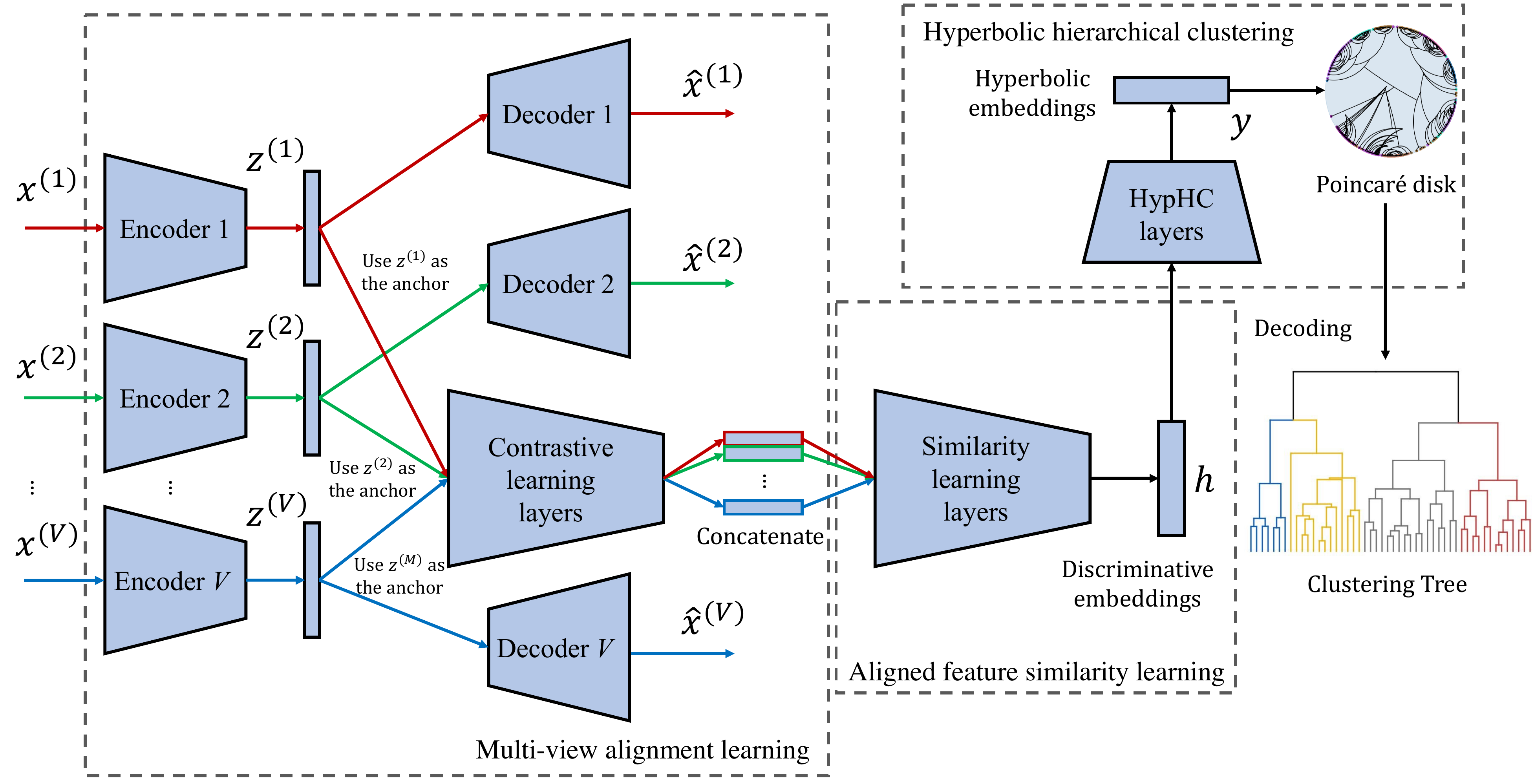}
	\caption{Overview of CMHHC. $V$ different autoencoders are assigned for $V$ different views. Contrastive learning layers are to align sample-level representations across multiple views. Similarity learning layers are to learn better metric property over aligned features. Then, HypHC layers are to optimize the tree-like embeddings in hyperbolic space. Finally, optimal hyperbolic embeddings are decoding into discrete clustering trees.}
	\label{fig:framework}
\end{figure}

Compared with partitional clustering and single-view hierarchical clustering, multi-view hierarchical clustering is less investigated. It requires a finer-grained understanding of both the consistency and differences among multiple views.
To this end, we propose a novel Contrastive Multi-view Hyperbolic Hierarchical Clustering~(CMHHC) model, as depicted in Figure~\ref{fig:framework}. 
The proposed model consists of three components, i.e., multi-view alignment learning, aligned feature similarity learning, and continuous hyperbolic hierarchical clustering.
First, to encourage consistency among multiple views and to suppress view-specific noise, we align the representations by contrastive learning, whose intuition is that the same instance from different views should be mapped closer while different instances should be mapped separately.
Next, to improve the metric property among data instances, we exploit more reasonable similarity of aligned features measured both on manifolds and in the Euclidean space, upon which we mine hard positives and hard negatives for unsupervised metric learning.
Besides, we also assign autoencoders to each view to regularize the model and prevent model collapse.
Then, we embed the representations with good similarities into the hyperbolic space and optimize the hyperbolic embeddings via the continuous relaxation of Dasgupta’s discrete objective for HC~\cite{chami2020trees}. 
Finally, the tree decoding algorithm helps decode the binary clustering tree from optimized hyperbolic embedding with low distortion.

To the best of our knowledge, the only relevant work on hierarchical clustering for multi-view data is the multi-view hierarchical clustering~(MHC) model proposed by Zheng \emph{et al.}~\shortcite{zheng2020multi}. MHC clusters multi-view data by alternating the cosine distance integration and the nearest neighbor agglomeration. 
Compared with MHC and na\"ive methods like multi-view concatenation followed by single-view hierarchical agglomerative clustering, our method enjoys several advantages.
Firstly, compared with simply concatenating multiple views in na\"ive methods or the averaging operation in MHC, CMHHC incorporates a contrastive multi-view alignment process, which can better utilize the complementary and consensus information among different views to learn meaningful view-invariance features of instances.
Secondly, compared with the shallow HC framework, deep representation learning and similarity learning are applied in our model to match complex real-world multi-view datasets and obtain more discriminative representations.
Thirdly, compared with heuristic agglomerative clustering, our method is oriented to gradient-based clustering process by optimizing multi-view representation learning and hyperbolic hierarchical clustering loss functions. 
These loss functions provide explicit evidence for measuring the quality of tree structure, which is crucial to achieving better HC performance.

The contributions of this work are summarized as follows:
\begin{itemize}
\item To our knowledge, we propose the first deep neural network model for multi-view hierarchical clustering, which can capture the aligned and discriminative representations across multiple views and perform hierarchical clustering at diverse levels of granularity.
\item To learn representative and discriminative multi-view embeddings, we exploit a contrastive representation learning module (to align representations across multiple views) and an aligned feature similarity learning module (to consider the manifold similarity). These embeddings are helpful to the downstream task of similarity-based clustering.
\item We validate our framework with five multi-view datasets and demonstrate that CMHHC outperforms existing HC and MHC algorithms in terms of the Dendrogram Purity~(DP) measurement.
\end{itemize}

\section{Related Work}
We review related work from three perspectives: hierarchical clustering, multi-view clustering, and hyperbolic geometry. 

\subsection{Hierarchical Clustering}
Hierarchical clustering raises a recursive way for partitioning a dataset into successively finer clusters.
Classical heuristics, like Single Linkage, Complete Linkage, Average Linkage, and Ward Linkage, are often the methods of choice for small datasets but may not scale well to large datasets as the running time scales cubically with the sample size~\cite{kobren2017hierarchical}.
Another problem with these heuristics is the lack of good objective functions, so there is no solid theoretical basis to support HC algorithms.
To overcome the challenge, Dasgupta~\shortcite{dasgupta2016cost} defined a proper discrete hierarchical clustering loss on all possible hierarchies.
Recently, gradient-based HC has gained increasing research attention, like feature-based gHHC~\cite{monath2019gradient}, and similarity-based UFit~\cite{chierchia2020ultrametric} and HypHC~\cite{chami2020trees}.

This paper deals with multi-view hierarchical clustering. The most relevant work to this paper is MHC~\cite{zheng2020multi}, which performs the cosine distance integration and the nearest neighbor agglomeration alternately.
However, this shallow HC framework ignores extracting meaningful and consistent information from multiple views, probably resulting in degenerated clustering performance.

\subsection{Multi-view Clustering}

Existing multi-view clustering~(MVC) methods mainly focus on partitional clustering.
Traditional multi-view clustering includes four types, i.e., multi-view subspace clustering~\cite{li2019reciprocal}, multi-view spectral clustering~\cite{kang2020multi}, multi-view matrix factorization-based clustering~\cite{cai2013multi}, and canonical correlation analysis~(CCA)-based clustering~\cite{chaudhuri2009multi}.
However, many MVC methods do not meet the requirements for complex nonlinear situations. Thus, deep MVC approaches have been attached recently~\cite{trosten2021reconsidering,xu2021multi,Xu_2022_CVPR}.
For example, Andrew \emph{et al.}~\shortcite{andrew2013deep} and Wang \emph{et al.}~\shortcite{wang2015deep} proposed deep versions of CCA, termed as deep CCA and deep canonically correlated autoencoders respectively.
Also, End-to-end Adversarial-attention network for Multi-modal Clustering~(EAMC)~\cite{zhou2020end} leveraged adversarial learning and attention mechanism to achieve the separation and compactness of cluster structure.

Despite many works towards partitional clustering, there is minimal research towards multi-view hierarchical clustering, which we address in this paper.

\subsection{Hyperbolic Geometry}
Hyperbolic geometry is a non-Euclidean geometry with a constant negative curvature, which drops the parallel line postulate of the postulates of Euclidean geometry~\cite{sala2018representation}. Since the surface area lying in hyperbolic space grows exponentially with its radius, hyperbolic space can be seen as a continuous version of trees whose number of leaf nodes also increases exponentially with the depth~\cite{monath2019gradient}. 
Hyperbolic geometry has been adopted to fertilize research related to tree structures recently~\cite{chami2020trees,nickel2017poincare,yan2021unsupervised}.


\section{Method}
Given a set of $N$ data points including $K$ clusters and $V$ different views $\left\{\boldsymbol{x}_i^1,\boldsymbol{x}_i^2,\cdots,\boldsymbol{x}_i^V\right\}_{i=1}^N$, where $\boldsymbol{x}_i^v\in\mathbb{R}^{D_v}$ denotes the $i$-th $D_v$-dimensional instance from the~$v$-th view,
we aim to produce a nested series of partitions, where the more similar multi-view samples are grouped earlier and have lower lowest common ancestors~(LCAs).
To this end, we establish a multi-view hierarchical clustering framework named CMHHC.
We first introduce the network architecture, and then we define the loss functions and introduce the optimization process.

\subsection{Network Architecture}
The proposed network architecture consists of a multi-view alignment learning module, an aligned feature similarity learning module, and a hyperbolic hierarchical clustering module, which is shown in Figure~\ref{fig:framework}. We introduce the details of each component as follows.

\subsubsection{Multi-view Alignment Learning}
Data from different sources tend to contain complementary and consensus information, so how to extract comprehensive representations and suppress the view-specific noise is the major task in multi-view representation learning. 
Therefore, we design the multi-view alignment learning module to map the data of each view into a low-dimentional aligned space. 
First of all, we assign a deep autoencoder~\cite{hinton2006reducing} to each view.
The reconstruction from input to output not only helps each view keep the view-specific information to prevent model collapse, but also fertilizes subsequent similarity computing through representation learning.
Specifically, considering the~$v$-th view $\boldsymbol{X}^{(v)}$, the corresponding encoder is represented as $\boldsymbol{Z}^{(v)} = E^{(v)}(\boldsymbol{X}^{(v)};\theta^{(v)}_\textrm e)$, and the decoder is represented as $\Hat{\boldsymbol{X}}^{(v)} = D^{(v)}(\boldsymbol{Z}^{(v)};\theta^{(v)}_\textrm d)$, where~$\theta^{(v)}_\textrm e$ and $\theta^{(v)}_\textrm d$ denote the autoencoder network parameters, and~$\boldsymbol{Z}^{(v)} \in\mathbb{R}^{N\times D_{\textrm{ae}}}$ and~$\Hat{\boldsymbol{X}}^{(v)}\in\mathbb{R}^{N\times D_{v}}$ denote learned~$D_{\textrm{ae}}$-dimensional latent features and the reconstructed output, respectively.

For the still-detached multiple views, inspired by recent works with contrastive learning~\cite{Xu_2022_CVPR}, we propose to achieve the consistency of high-level semantic features across views in a contrastive way, i.e., corresponding instances of different views should be mapped closer, while different instances should be mapped more separately.
To be specific, we define a cross-view positive pair as two views describing the same object, and a negative pair as two different objects from the same view or two arbitrary views.
We encode the latent features to aligned features denoted as~$\boldsymbol{H}^{(v)} = f_{\textrm{con}}(\boldsymbol{Z}^{(v)};\theta_\textrm c)$, where~$\theta_\textrm c$~denotes the parameters of contrasive learning encoder~$f_{\textrm{con}}(\cdot)$, and $\boldsymbol{H}^{(v)}\in\mathbb{R}^{N\times D_\textrm h}$.
Then, we compute the cosine similarity~\cite{chen2020simple} of two representations $\boldsymbol{h}_{i}^{(v_1)}$ and $\boldsymbol{h}_{j}^{(v_2)}$($v_1\neq v_2$):
\begin{equation}
\small
d_{i,j}^{(v_1)(v_2)} = \frac{\boldsymbol{h}_{i}^{(v_1)T}\boldsymbol{h}_{j}^{(v_2)}}{\left\|\boldsymbol{h}_{i}^{(v_1)}\right\| \cdot \left\|\boldsymbol{h}_{j}^{(v_2)}\right\|}\,.
\end{equation}
To achieve consistency of all views, we expect the similarity scores of positive pairs~$(\boldsymbol{h}_{i}^{(v_1)}, \boldsymbol{h}_{i}^{(v_2)})$ to be larger, and those of negative pairs~$(\boldsymbol{h}_{i}^{(v_1)}, \boldsymbol{h}_{j}^{(v_2)})$ and $(\boldsymbol{h}_{i}^{(v)}, \boldsymbol{h}_{j}^{(v)})$ to be smaller. 

\subsubsection{Aligned Feature Similarity Learning}

After the multi-view alignment learning, comprehensive representations of multi-view data are obtained, and view-specific noise is suppressed.
However, the aligned hidden representations do not explicitly guarantee an appropriate similarity measurement that preserves desired distance structure between pairs of multi-view instances, which is crucial to similarity-based hierarchical clustering. Therefore, we devise an aligned feature similarity learning module to optimize the metric property of similarity used in hierarchy learning.

Intuitively, samples may be mapped to some manifold embedded in a high dimensional Euclidean space, so it would be beneficial to utilize the distance on manifolds to improve the similarity measurement~\cite{iscen2018mining}.
This module performs unsupervised metric learning by mining positives and negatives pairs with both the manifold and Euclidean similarities.
Similar with Iscen \emph{et al.}~\shortcite{iscen2018mining}, we measure the Euclidean similarity 
of any sample pair via the mapping function~$w^{\textrm e}_{i,j}(\boldsymbol{h}_i,\boldsymbol{h}_j) = \max(0,\boldsymbol{h}_i^T\boldsymbol{h}_j)^3$ and refer to $\textrm{ENN}_k(\boldsymbol{h}_i)$ as Euclidean $k$ nearest neighbors~($k$NN) set of~$\boldsymbol{h}_i$, where~$\boldsymbol{h}_i\in\boldsymbol{H}$ is the concatenate aligned representation.

In terms of the manifold similarity, the Euclidean affinity graph needs to be calculated as preparations. The elements of the affinity matrix~$A$ are weighted as~$a_{ij} = w^{\textrm e}_{i,j}(\boldsymbol{h}_i,\boldsymbol{h}_j)$ when~$\boldsymbol{h}_i$ and $\boldsymbol{h}_j$ are both the Euclidean $k$NN nodes to each other, or else~$a_{ij} = 0$. 
Following the spirit of advanced random walk model~\cite{iscen2017efficient}, we can get the convergence solution~$\boldsymbol{r}_i$ efficiently, where an element  $\boldsymbol{r}_i(j)$ denotes the ``best''  walker from the~$i$-th node to the~$j$-th node. Therefore, the manifold similarity function can be defined as $w^{\textrm m}_{i,j}(\boldsymbol{h}_i,\boldsymbol{h}_j) = \boldsymbol{r}_i(j)$. Similarly, we denote $\textrm{MNN}_k(\boldsymbol{h}_i)$ as the manifold~$k$NN set of~$\boldsymbol{h}_i$. 

To this end, we try to consider every data point in the dataset as an anchor in turn so that the hard mining strategy keeps feasible under the premise of acceptable computability. Given an anchor $\boldsymbol{h}_i$ from $\boldsymbol{H}$, we select~$k_{\textrm{pos}}$ nearest feature vectors on manifolds, which are not that close in the Euclidean space, as good positives. By $\textrm{ENN}_{k_{\textrm{pos}}}(\boldsymbol{h}_i)$ and $\textrm{MNN}_{k_{\textrm{pos}}}(\boldsymbol{h}_i)$, the hard positive set is descendingly sorted by the manifold similarity as:
\begin{equation}
\small
S_{\textrm{pos}}(\boldsymbol{h}_i) = \textrm{MNN}_{k_{\textrm{pos}}}(\boldsymbol{h}_i) - \textrm{ENN}_{k_{\textrm{pos}}}(\boldsymbol{h}_i)\,,
\end{equation} 
where~$k_{\textrm{pos}}$ decides how hard the selected positives are, which is a completely detached value from $k$. However, to keep gained pseudo-label information with little noise, good negatives are expected to be not only relatively far from the anchor on manifolds but also in the Euclidean space. So the hard negative set is in the descending order according to the Euclidean similarity, denoted as:
\begin{equation}
\small
S_{\textrm{neg}}(\boldsymbol{h}_i) = S_{\textrm{all}}(\boldsymbol{H}) - (\textrm{ENN}_{k_{\textrm{neg}}}(\boldsymbol{h}_i) + \textrm{MNN}_{k_{\textrm{neg}}}(\boldsymbol{h}_i))\,,
 \end{equation}
where $S_{\textrm{all}}(\boldsymbol{H})$ is the set of all feature vectors, and $k_{\textrm{neg}}$ is the value of the nearest neighbors for negatives, separated from~$k$. 
Intuitively, a larger value of~$k_{\textrm{pos}}$ leads to harder positives for considering those with relatively lower confidence, and tolerating the intra-cluster variability. Similarly, a smaller value of~$k_{\textrm{neg}}$ means harder negatives for distinguishing the anchors from more easily-confused negatives. 

After obtaining hard tuples as the supervision of similarity learning for multi-view input, we are able to get clustering-friendly embeddings on top of the aligned feature space via an encoder $\boldsymbol{E} = f_{\textrm{mom}}(\boldsymbol{H};\theta_\textrm m)$, where $\boldsymbol{E}\in\mathbb{R}^{N\times D_\textrm e}$ is the discriminative embeddings of $D_\textrm e$ dimensions and $\theta_\textrm m$ is the parameter of the encoder $f_{\textrm{mom}}(\cdot)$.

\subsubsection{Hyperbolic Hierarchical Clustering}
To reach the goal of hierarchical clustering, we adopt the continuous~optimizing process of~Hyperbolic Hierarchical Clustering~\cite{chami2020trees} as the guidance of this module, which is stacked on the learned common-view embeddings. By means of an embedder~$f_{\textrm{hc}}(\cdot)$ parameterized by~$\theta_{\textrm{hc}}$, we embed the learned similarity graph from common-view space to a specific~hyperbolic space, i.e., the Poincar\'{e} Model~$\mathbb{B}_h = {\left\{\left\|\boldsymbol{y}\right\|_2\leq 1,\boldsymbol{y}\in\mathbb{R}^h\right\}}$ whose curvature is constant~$-1$.
The distance of the geodesic between any two points~$\boldsymbol{y}_i$,~$\boldsymbol{y}_j\in\mathbb{B}_h$ in hyperbolic space is~$d_{\mathbb{B}}(\boldsymbol{y}_i,\boldsymbol{y}_j)=\rm cosh^{-1}(1+\frac{2\left\|\boldsymbol{y}_i-\boldsymbol{y}_j\right\|_2^2}{(1-{\left\|\boldsymbol{y}_i\right\|}_2^2)(1-{\left\|\boldsymbol{y}_i\right\|}_2^2)})$. With this prior knowledge, we can find the optimal hyperbolic embeddings~$\boldsymbol{Y}^{*}$ pushed to the boundary of the ball via a relaxed form of improved~Dasgupta's cost function. Moreover, the hyperbolic LCA of two hyperbolic embeddings~$\boldsymbol{y}_i$ and~$\boldsymbol{y}_j$ is the point on their geodesic nearest to the origin of the ball. It can be represented as~$\boldsymbol{y_i}\vee\boldsymbol{y_j}:=\mathop{\arg\min}_{\boldsymbol{y}_o\in\boldsymbol{y}_i\leadsto\boldsymbol{y}_j}d(\boldsymbol{o},\boldsymbol{y})$. The LCA of two hyperbolic embeddings is an analogy to that of two leaf nodes~$\boldsymbol{t}_i$ and~$\boldsymbol{t}_j$ in a discrete tree, where the tree-like~LCA is the node closest to the root node on the two points' shortest path~\cite{chami2020trees}. Based on this, the best hyperbolic embeddings can be decoded back to the original tree via getting merged iteratively along the direction of the ball radius from boundary to origin. 

\subsection{Loss Functions and Optimization Process}
This section introduces the loss functions for CMHHC, including multi-view representation learning loss and hierarchical clustering loss, and discusses the optimization process.

\subsubsection{Multi-view Representation Learning Loss}
In our model, we jointly align multi-view representations and learn the reliable similarities in common-view space by integrating the autoencoder reconstruction loss $\mathcal{L}_\textrm r$, the multi-view contrastive loss $\mathcal{L}_\textrm c$ and the positive weighted triplet metric loss $\mathcal{L}_\textrm m$, so the objective for multi-view representation learning loss $\mathcal{L}_{\textrm{mv}}$ is defined as:
\begin{equation}
\label{eq:mvrl_loss}
\small
\mathcal{L}_{\textrm{mv}}=\mathcal{L}_\textrm r+\mathcal{L}_\textrm c+\mathcal{L}_\textrm m\,.
\end{equation}
First of all, $\mathcal{L}_\textrm r^{(v)}$ is the~$v$-th view loss function of reconstructing~$\boldsymbol{x}_i^{(v)}$, so the complete reconstruction loss for all views is:
\begin{equation}
\small
\label{eq:ae}
\mathcal{L}_\textrm r = \sum\limits_{v=1}^{V}\mathcal{L}_\textrm r^{(v)} = \sum\limits_{v=1}^{V}\frac{1}{N}\sum\limits_{i=1}^{N}{\left\|\boldsymbol{x}_i^{(v)}-D^{(v)}(E^{(v)}(\boldsymbol{x}_i^{(v)}))\right\|}_2^2\,.
\end{equation}
As for the second term, assuming that alignment between every two views ensures alignment among all views, we introduce total multi-view mode of contrastive loss as follows:
\begin{equation}
\label{eq:con}
\small
\mathcal{L}_\textrm c = \sum\limits_{v_1=1}^{V}\sum\limits_{v_2=1,v_2\neq v_1}^{V}\mathcal{L}_\textrm c^{(v_1)(v_2)}\,,
\end{equation}
where the contrastive loss function between reference view~$v_1$ and contrast view~$v_2$ is:
\begin{equation}
\label{eq:each_con}
\small
\mathcal{L}_\textrm c^{(v_1)(v_2)} = -\frac{1}{N}\sum\limits_{i=1}^{N}\log\frac{e^{d_{i,i}^{(v_1)(v_2)}/\tau}}{\sum\limits_{j=1,j\neq i}^{N}e^{d_{i,j}^{(v_1)(v_1)}/\tau}+\sum\limits_{j=1}^{N}e^{d_{i,j}^{(v_1)(v_2)}/\tau}}\,,
\end{equation}
where $\tau$ is the temperature hyperparameter for multi-view contrastive loss. 

The third term is for similarity learning. Hard tuples includes an anchor~$\boldsymbol{h}_i\in\boldsymbol{H}$, a hard positive sample~$\boldsymbol{h}_i^{\textrm{pos}}\in S_{\textrm{pos}}(\boldsymbol{h}_i)$ and a hard negative sample~$\boldsymbol{h}_i^{\textrm{neg}}\in S_{\textrm{neg}}(\boldsymbol{h}_i)$. Both the single~$\boldsymbol{h}_i^{\textrm{pos}}$ and the single~$\boldsymbol{h}_i^{\textrm{neg}}$ are randomly selected from corresponding sets. Then we calculate the embeddings of the tuple:~$\boldsymbol{e}_i=f_{\textrm{mom}}(\boldsymbol{h}_i;\theta_\textrm m)$,~$\boldsymbol{e}_i^{\textrm{pos}}=f_{\textrm{mom}}(\boldsymbol{h}_i^{\textrm{pos}};\theta_\textrm m)$ and~$\boldsymbol{e}_i^{\textrm{neg}}=f_{\textrm{mom}}(\boldsymbol{h}_i^{\textrm{neg}};\theta_\textrm m)$, so that we can measure the similarities of the embeddings via the weighted triplet loss:
\begin{equation}
\label{eq:metrc_learning}
\small
\mathcal{L}_{\textrm m} = \frac{1}{N}{\sum\limits_{i=1}^N} w^{\textrm m}(\boldsymbol{e}_i,\boldsymbol{e}_i^{\textrm{pos}})[m+{\left\|\boldsymbol{e}_i-\boldsymbol{e}_i^{\textrm{pos}}\right\|}_2^2-{\left\|\boldsymbol{e}_i-\boldsymbol{e}_i^{\textrm{neg}}\right\|}_2^2]_+\,,
\end{equation}
where~$w^{\textrm m}(\boldsymbol{e}_i,\boldsymbol{e}_i^{\textrm{pos}})$ represents the degree of contribution of every tuple. Weighting the standard triplet loss by the similarity between the anchor and the positive on manifolds relieves the pressure from the tuples with too hard positives.

It is worth mentioning that $\mathcal{L}_{\textrm{mv}}$ in Eq~(\ref{eq:mvrl_loss}) is optimized by mini-batch training so that the method can scale up to large multi-view datasets.

\subsubsection{Hierarchical Clustering Loss}
A ``good'' hierarchical tree means that more similar data instances should be merged earlier.
Dasgupta~\shortcite{dasgupta2016cost} first proposed an explicit hierarchical clustering loss function:
\begin{equation}
\small
\mathcal{L}_{\textrm{Dasgupta}}(T;w_{ij})) = \sum\limits_{ij}w^{\textrm e}_{i,j}\left|\textrm{leaves}(T[i\vee j])\right|\,,
\label{dasgupta}
\end{equation}
where $w^{\textrm e}_{i,j}$ is the Euclidean similarity between~$i$ and~$j$ and~$T[i\vee j]$ is the subtree rooted at the LCA of the~$i$ and~$j$ nodes, and~${\rm leaves}(T[i\vee j])$ denotes the set of descendant leaves of internal node~$T[i\vee j]$. 

Here, we adopt the differentiable relaxation of constrained Dasguta's objective~\cite{chami2020trees}.
Our pairwise similarity graph is learned from the multi-view representation learning process so that the unified similarity measurement narrows the gap between representation learning and clustering.
The hyperbolic hierarchical clustering loss for our model is defined as:
\begin{equation}
\label{eq:hyphc_loss}
\begin{aligned}
\small
 \mathcal{L}_{\textrm{hc}}(\boldsymbol{Y};w^{\textrm e},\tau_{\textrm c}) = \sum\limits_{i,j,k}(&w_{i,j}^{\textrm e}+w_{i,k}^{\textrm e}+w_{j,k}^{\textrm e}\\
 &-w_{i,j,k}^{\textrm{hyp}}(\boldsymbol{Y};w,\tau_{\textrm  c})+\sum\limits_{i,j}(w_{i,j}^{\textrm e})\,.
\end{aligned}
\end{equation}
We compute the hierarchy of any embedding triplet through the similarities of all embedding pairs among three samples:
\begin{equation}
\begin{aligned}
\small
w_{i,j,k}^{\textrm{hyp}}(\boldsymbol{Y};w^{\textrm e},\tau_{\textrm  c})=(w_{i,j}^{\textrm e},w_{i,k}^{\textrm e},w_{j,k}^{\textrm e})\cdot\sigma_{\tau_\textrm c}(d_{\mathbb{B}}(o,\boldsymbol{y}_i\vee\boldsymbol{y}_j),\\
d_{\mathbb{B}}(o,\boldsymbol{y}_i\vee\boldsymbol{y}_k),d_{\mathbb{B}}(o,\boldsymbol{y}_j\vee\boldsymbol{y}_k))^T\,,
\end{aligned}
\end{equation}
where~$\sigma_{\tau_{\textrm c}}(\cdot)$ is the softmax function~$\sigma_{\tau_\textrm  c}(d)_i=e^{d_i/\tau_{\textrm c}}/\sum_je^{d_i/\tau_{\textrm c}}$ scaled by the temperature parameter~$\tau_{\textrm c}$ for the hyperbolic hierarchical clustering loss. Eq.~(\ref{eq:hyphc_loss}) theoretically asks for similarities among all tuples of the dataset, which takes a high time complexity of~$O(N^3)$. However, we could sample $N^2$-order triplets by obtaining all possible node pairs and then choosing the third node randomly from the rest~\cite{chami2020trees}. With mini-batch training and sampling tricks, HC can scale to large datasets with acceptable computation complexity.

Then the optimal hyperbolic embeddings are denoted as:
\begin{equation}
\begin{aligned}
\small
\boldsymbol{Y}^*=\mathop{\arg\min}_{\boldsymbol{Y}}\mathcal{L}_{\textrm{hc}}(\boldsymbol{Y};w^{\textrm e},\tau_{\textrm c})\,.
\end{aligned}
\end{equation}
Finally, the nature of negative curvature and bent tree-like geodesics in hyperbolic space allows us to decode the binary tree~$T$ in the original space by grouping two similar embeddings whose hyperbolic LCA is the farthest to the origin:
\begin{equation}
\label{eq:decode}
\small
T = \rm dec(\boldsymbol{Y}^*)\,,
\end{equation}
where~$\rm dec(\cdot)$ is the decoding function~\cite{chami2020trees}.

\subsubsection{Optimization Process}

The optimization process is summarized in the Appendix. 
The entire training process of our framework has two steps: (A)~Optimizing the multi-view representation learning loss with Eq.~(\ref{eq:mvrl_loss}), and (B)~Optimizing the hyperbolic hierarchical clustering loss with Eq.~(\ref{eq:hyphc_loss}) .

\section{Experiments}

\subsection{Experimental Setup}

\subsubsection{Datasets}
We conduct our experiments on the following five real-world multi-view datasets.

\begin{itemize}
\item MNIST-USPS~\cite{peng2019comic} is a two-view dataset with 5000 hand-written digital~(0-9) images. The MNIST view is in $28\times 28$ size, and the USPS view is in $16\times 16$ size. 
\item BDGP~\cite{li2019deep} contains 2500 images of Drosophila embryos divided into five categories with two extracted features. One view is 1750-dim visual features, and the other view is 79-dim textual features.
\item Caltech101-7~\cite{dueck2007non} is established with 5 diverse feature descriptors, including 40-dim wavelet moments~(WM), 254-dim CENTRIST, 1,984-dim HOG, 512-dim GIST, and 928-dim LBP features, with 1400 RGB images sampled from 7 categories. 
\item COIL-20 contains object images of 20 categories. Following Trosten \emph{et al.}~\shortcite{trosten2021reconsidering}, we establish a variant of COIL-20, where 480 grayscale images of $128 \times 128$ pixel size are depicted from 3 different random angles.
\item Multi-Fashion~\cite{Xu_2022_CVPR} is a three-view dataset with 10,000 $28\times 28$ images of different fashionable designs, where different views of each sample are different products from the same category.
\end{itemize}


\subsubsection{Baseline Methods}
We demonstrate the effects of our CMHHC by comparing with the following three kinds of hierarchical clustering methods. Note that for single-view methods, we concatenate all views into a single view to provide complete information~\cite{peng2019comic}.

Firstly, we compare CMHHC with conventional linkage-based discrete single-view hierarchical agglomerative clustering~(HAC) methods, including Single-linkage, Complete-linkage, Average-linkage, and Ward-linkage algorithms. 
Secondly, we compare CMHHC with the most representative similarity-based continuous single-view hierarchical clustering methods, i.e., UFit~\cite{chierchia2020ultrametric}, and HypHC~\cite{chami2020trees}.
For UFit, we adopt the best loss function~``\emph{Closest+Size}''.
As for HypHC, we directly use the continuous Dasgupta's objective.
We utilized the corresponding open-source versions for both of the methods and followed the default parameters in the codes provided by the authors.
Lastly, to the best of our knowledge, MHC~\cite{zheng2020multi} is the only proposed multi-view hierarchical clustering method.
We implemented it with Python as there was no open-source implementation available.

\subsubsection{Hierarchical Clustering Metrics}
Unlike partitional clustering, binary clustering trees, i.e., the results of hierarchical clustering, can provide diverse-granularity cluster information when the trees are truncated at different altitudes. Hence, general clustering metrics, such as clustering accuracy~(ACC) and Normalized Mutual Information~(NMI), is not able to display the characteristics of clustering hierarchies comprehensively.
To this end, following previous literature~\cite{kobren2017hierarchical,monath2019gradient}, we validate hierarchical clustering performance via the Dendrogram Purity~(DP) measurement.
To sum up, DP measurement is equivalent to the average purity over the leaf nodes of LCA of all available data point pairs in the same ground-truth clusters. Clustering tree with higher DP value contains purer subtrees and keeps a more consistent structure with ground-truth flat partitions. 
A detailed explanation of DP is in the appendix. 

\subsubsection{Implementation Details}
Our entire CMHHC model is implemented with PyTorch. 
We first pretrain $V$ autoencoders for 200 epochs and contrastive learning encoder for 10, 50, 50, 50 and 100 epochs on BDGP, MNIST-USPS, Caltech101-7, COIL-20, and Multi-Fashion respectively, and then finetune the whole multi-view representation learning process for 50 epochs, and finally train the hyperbolic hierarchical clustering loss for 50 epochs. 
The batch size is set to 256 for representation learning and 512 for hierarchical clustering, using the Adam and hyperbolic-matched Riemannian optimizer~\cite{kochurov2020geoopt} respectively. 
The learning rate is set to~$5e^{-4}$ for Adam, and a search over~$[5e^{-4}, 1e^{-3}]$ for Riemannian Adam of different datasets.
We empirically set~$\tau=0.5$ for all datasets, while~$\tau_{\textrm c}=5e^{-2}$ for BDGP, MNIST-USPS, and Multi-Fashion and $\tau_\textrm c=1e^{-1}$ for Caltech101-7 and COIL-20. 
We run the model 5 times and report the results with the lowest value of~$L_{\textrm c}$.
In addition, we create an adjacency graph with 50 Euclidean nearest neighbors to compute manifold similarities.
We make the general rule that the~$k_{\textrm{pos}}$ value equals $N/2K$ and the~$k_{\textrm{neg}}$ value equals $N/K$, making the selected tuples hard and reliable. 
More detailed parameter setting can be found in the Appendix.

\subsection{Experimental Results}
\subsubsection{Performance Comparison with Baselines}
Experimental DP results are reported in Table~\ref{tab:result}. The results illustrate that our unified CMHHC model outperforms comparing baseline methods. As it shows, our model gains a significant growth in DP by~$11.47\%$ on BDGP,~$1.61\%$ on MNIST-USPS,~$20.00\%$ on Caltech101-7,~$4.08\%$ on COIL-20 and~$23.92\%$ on Multi-Fashion over the second-best method. The underlying reason is that~CMHHC captures much more meaningful multi-view aligned embeddings instead of concatenating all views roughly without making full use of the complementary and consensus information. Our deep model greatly exceeds the level of the only multi-view hierarchical clustering work MHC, especially on Caltech101-7, COIL-20, and large-scale Multi-Fashion. This result can be attributed to the alignment and discrimination of the multi-view similarity graph learning for hyperbolic hierarchical clustering. Additionally, the performance gap between our model and deep continuous UFit and HypHC reflects the limitations of fixing input graphs without an effective similarity learning process.

\begin{table}[t]
\renewcommand{\arraystretch}{\tabelarraystretch}
    \centering
    \resizebox{\columnwidth}{!}{
    \begin{tabular}{c|c|c|c|c|c}
        \toprule
        Method & MNIST-USPS & BDGP & Caltech101-7 & COIL-20 & Multi-Fashion\\
        \midrule
        HAC (Single-linkage)   & 29.81\% & 61.88\% & 23.67\% & 72.56\% & 27.89\% \\
        HAC (Complete-linkage) & 54.36\% & 56.57\% & 30.19\% & 69.95\% & 48.72\% \\
        HAC (Average-linkage)  & 69.67\% & 45.91\% & 30.90\% & 73.14\% & 65.70\% \\
        HAC (Ward-linkage)     & \underline{80.38\%} & 58.61\% & 35.69\% & \underline{80.81\%} & \underline{72.33\%} \\
        \midrule
        UFit                   & 21.67\% & 69.20\% & 19.00\% & 55.41\% & 25.94\% \\
        HyperHC                & 32.99\% & 31.21\% & 22.46\% & 28.50\% & 25.65\% \\
        \midrule
        MHC                    & 78.27\% & \underline{89.14\%} & \underline{45.22\%} & 66.50\% & 54.81\% \\
        CMHHC (Ours)           & \textbf{94.49\%} & \textbf{91.53\%} & \textbf{66.52\%} & \textbf{84.89\%} & \textbf{96.25\%} \\
        \bottomrule
    \end{tabular}
    }
    \caption{Dendrogram Purity~(DP) results of baselines and CMHHC. We concatenate the features of multi-view data for single-view baselines, including HAC, UFit, and HyperHC.}
    \label{tab:result}
\end{table}

\subsubsection{Ablation Study}
We conduct an ablation study to evaluate the effects of the components in the multi-view representation learning module. To be specific, we refer to the~CMHHC without autoencoders for reconstruction of multiple views as CMHHC$_{\neg \textrm{AE}}$, without contrastive learning submodel as CMHHC$_{\neg \textrm{Con}}$, and without similarity learning as CMHHC$_{\neg \textrm{Sim}}$. 
We train CMHHC$_{\neg \textrm{AE}}$, CMHHC$_{\neg \textrm{Con}}$ and CMHHC$_{\neg \textrm{Sim}}$ after removing the corresponding network layers. Table~\ref{tab:ablation} shows the experimental results on 5 datasets.
The results show that the proposed components contribute to the final hierarchical clustering performance in almost all cases.

\begin{table}[t]
\renewcommand{\arraystretch}{\tabelarraystretch}
    \centering
    \resizebox{0.925\columnwidth}{!}{
    \begin{tabular}{c|c|c|c|c|c}
        \toprule
        Ablation & MNIST-USPS & BDGP & Caltech101-7 & COIL-20 & Multi-Fashion\\
        \midrule
        CMHHC           & \textbf{94.49\%} & \textbf{91.53\%} & 66.52\% & \textbf{84.89\%} & \textbf{96.25\%} \\
        \midrule
        CMHHC$_{\neg \textrm{AE}}$   & 92.92\% & 86.19\% & \textbf{68.50\%} & 27.16\% & 89.06\% \\
        CMHHC$_{\neg \textrm{Con}}$  & 43.10\% & 24.51\% & 33.02\% & 14.73\% & 43.57\% \\
        CMHHC$_{\neg \textrm{Sim}}$  & 89.78\% & 90.10\% & 19.74\% & 53.40\% & 44.65\% \\
        \bottomrule
    \end{tabular}
    }
    \caption{Ablation study results.}
    \label{tab:ablation}
\end{table}

\subsubsection{Case Study}

\begin{figure}[t]
	\centering
	\includegraphics[width=\linewidth]{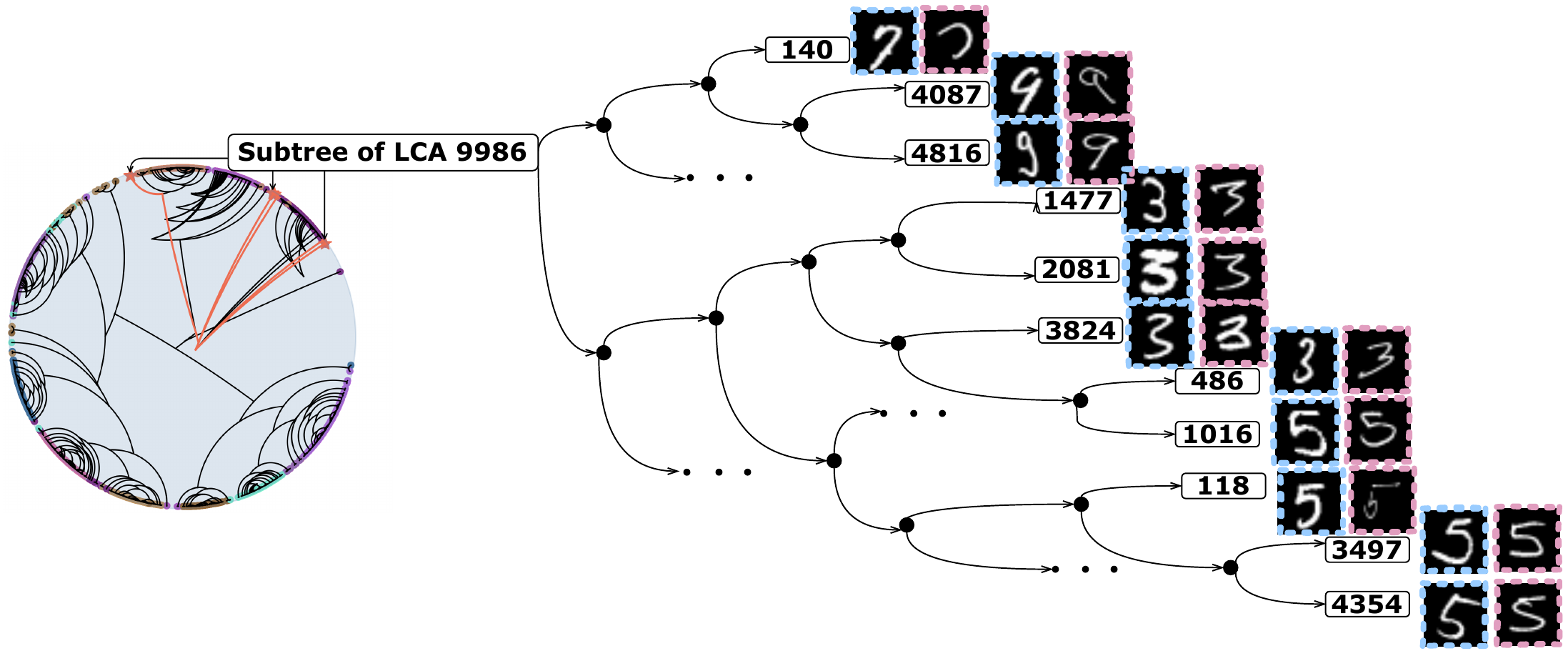}
	\caption{Visualization of a truncated subtree from the decoded tree on the MNIST-USPS dataset. The right part represents the sampled subtree structure of LCA \#$9986$, where MNIST images are framed in blue, and USPS images are framed in orange. We observe that more similar pairs will have lower LCAs. For example, images belonging to the same category (like \#$3497$ and \#$4354$ of digit $5$) are grouped together first, i.e., have the lowest LCA, while less similar images (like \#$3497$ of digit $5$ and \#$140$ of digit $7$) are merged at the highest LCA in the subtree.}
	\label{fig:case}
\end{figure}

We qualitatively evaluate a truncated subtree structure learned via our method for the hierarchical tree. We plot the sampled MNIST-USPS subtrees of the final clustering tree in Figure~\ref{fig:case}. As shown, the similarity between two nodes is getting more substantial from the root to the leaves, indicating that the hierarchical tree can reveal fine-grained similarity relationships and ground-truth flat partitions for the multi-view data.

\section{Conclusion}
This paper proposed a novel multi-view hierarchical clustering framework based on deep neural networks. Employing multiple autoencoders, contrastive multi-view alignment learning, and unsupervised similarity learning, we capture the invariance information across views and learn the meaningful metric property for similarity-based continuous hierarchical clustering. Our method aims at providing a clustering tree with high interpretability oriented towards multi-view data, highlighting the importance of representations' alignment and discrimination, and indicating the potential of gradient-based hyperbolic hierarchical clustering. Extensive experiments illustrate CMHHC is capable of clustering multi-view data at diverse levels of granularity.

\section*{Acknowledgments}
This work was partially supported by the National Key Research and Development Program of China (No. 2018AAA0100204), and a key  program of fundamental research from Shenzhen Science and Technology Innovation Commission (No. JCYJ20200109113403826).

\newpage
\bibliographystyle{named}
\bibliography{main}

\clearpage
\appendix
\section{Algorithm Pseudocode for CMHHC}
Algorithm~\ref{alg:cmhhc} presents the step-by-step procedure of the proposed CMHHC.
\begin{algorithm}[ht]
\caption{Pseudocode to optimize our CMHHC}
\label{alg:cmhhc}
\textbf{Input}: Multi-view dataset $\left\{\boldsymbol{x}_i^1,\boldsymbol{x}_i^2,\cdots,\boldsymbol{x}_i^V\right\}_{i=1}^N$;

\quad\quad\quad Hyperparameters $\tau$, $\tau_\textrm{c}$, $k$, $k_{\textrm{pos}}$ and $k_{\textrm{neg}}$;\\
\textbf{Output}: Final binary clustering tree.

\begin{algorithmic}[1] 
\STATE \textbf{Initialization}: Initialize the network parameters $\left\{\theta^{(v)}_\textrm{e}, \theta^{(v)}_\textrm{d}\right\}_{v=1}^V$, $\theta_\textrm{c}$, $\theta_\textrm{m}$ and $\theta_{\textrm{hc}}$.
\STATE\textbf{Pretrain}: Update $\left\{\theta^{(v)}_\textrm{e}, \theta^{(v)}_\textrm{d}\right\}_{v=1}^V$ by Eq. (\ref{eq:ae}).
\STATE\quad\quad\quad\quad Update $\theta_\textrm{c}$ by Eq. (\ref{eq:con}).
\STATE Performing hard mining strategy.
\STATE \textbf{Step(A)}: To learn $\boldsymbol{E}$.\\
\STATE\quad\quad\quad\quad Update~$\left\{\theta^{(v)}_\textrm{e}, \theta^{(v)}_\textrm{d}\right\}_{v=1}^V$,~$\theta_\textrm{c}$,~$\theta_\textrm{m}$ by Eq.~(\ref{eq:mvrl_loss}).
\STATE\textbf{Step(B)}: To learn $\boldsymbol{Y}^*$.
\STATE\quad\quad\quad\quad Update $\theta_{\textrm{hc}}$ by Eq. (\ref{eq:hyphc_loss}).
\STATE Decoding $T$ from $\boldsymbol{Y^*}=\left\{\boldsymbol{y_1^*}, \dots, \boldsymbol{y_N^*}\right\}$ by Eq.(\ref{eq:decode}).

\end{algorithmic}
\end{algorithm}

\section{Summarization of Notations}

\begin{table}[th]
    \centering
	\resizebox{\columnwidth}{!}{
    \begin{tabular}{r|p{9.35cm}}
		\toprule  
        Notation & Definition \\
        \midrule
        $\mathcal{X}$ & The set of training examples.\\
        $\boldsymbol{X}^{(v)}$ & The data samples in the $v$-th view.\\
        $\boldsymbol{Z}^{(v)}$ & The latent features of $v$-th view. \\
        $\Hat{\boldsymbol{X}}^{(v)}$ & The  reconstructed samples in the $v$-th view. \\ 
        $\boldsymbol{H}^{(v)}$ & The aligned representations of $v$-th view. \\
        $\boldsymbol{H}$ & The concatenate aligned representations among all views.\\
        $\boldsymbol{E}$ & The discriminative embeddings among all views. \\
        $\boldsymbol{Y}$ & The hyperbolic embeddings in Poincar\'{e} Model. \\
        $\boldsymbol{Y}^*$ & The optimal hyperbolic embeddings. \\
        $T$ & Binary HC decoding tree. \\
        $\boldsymbol{x}_i^{(v)}$ & The $i$-th data sample in the $v$-th view. \\
        $\boldsymbol{z}_i^{(v)}$ & The $i$-th latent feature in the $v$-th view. \\
        $\Hat{\boldsymbol{x}_i}^{(v)}$ & The $i$-th reconstructed sample in the $v$-th view. \\
        $\boldsymbol{h}_{i}^{(v)}$ & The $i$-th aligned representation in the $v$-th view.\\
        $\boldsymbol{h}_{i}$ & The $i$-th concatenate aligned representations.\\
        $\boldsymbol{h}_{i}^{\textrm{pos}}$ & Hard positive representation for the $i$-th anchor representation.\\
        $\boldsymbol{h}_{i}^{\textrm{neg}}$ & Hard negative representation for the $i$-th anchor representation.\\
        $\boldsymbol{e}_{i}$ & The $i$-th discriminative embedding from multiple views.\\
        $\boldsymbol{e}_{i}^{\textrm{pos}}$ & Hard positive embedding for the $i$-th anchor embedding.\\
        $\boldsymbol{e}_{i}^{\textrm{neg}}$ & Hard negative embedding for the $i$-th anchor embedding.\\
        $\boldsymbol{y}_i$ & The $i$-th hyperbolic embeddings. \\
        $N$ & The number of data instances (leaf nodes). \\
        $K$ & The number of clusters.\\
        $V$ & The number of views. \\
        $C$ & The number of clusters. \\
        $D_v$ & The dimensionality of the $v$-th view. \\
        $D_{\textrm{ae}}$ & The dimensionality of latent space. \\
        $D_\textrm{h}$ & The dimensionality of aligned representation space. \\
        $D_\textrm{e}$ & The dimensionality of discriminative embedding space. \\
        $d_{i,j}^{(v_1)(v_2)}$ & Cosinne distance of the $i$-th in the $v_1$-th view and the $j$-th in the $v_2$-th view in common-view space.\\
        $A$ & Affinity matrix of concatenate aligned representations.\\
        $a_{ij}$ & Adjacanncy of the $i$-th and $j$-th representations.\\
        $\boldsymbol{r}_i$ & The solution to random walk model of $i$-th sample.\\
        $\boldsymbol{r}_i(j)$ & The best walker from the $i$-th node to the $j$-th node.\\
        \bottomrule
    \end{tabular}
    }
    \caption{Notation used in CMHHC}
	\label{tb:notations}
\end{table}

We summarize the notations used in the paper in Table~\ref{tb:notations}.

\section{Experimental Details}

We introduce more experimental details in this section.
All the experiments are conducted on a Linux Server with TITAN Xp (10G) GPU and Intel(R) Xeon(R) CPU E5-2630 v4 @ 2.20GHz.

\subsection{Datasets}
We conduct our experiments on the following multi-view datasets. \textbf{MNIST-USPS}~\cite{peng2019comic} consists of 5000 hand-written digital~(0-9) images with 10 categories. The MNIST view is in $28\times 28$ size, sampled randomly from MNIST dataset, and the USPS view is in $16\times 16$ size, sampled randomly from USPS dataset. Each of category is with 500 images.
Specifically, for convenience of experiments, we adopt zero-padding to make the dimensions of USPS view $28\times 28$, the same as MNIST view.
\textbf{BDGP}~\cite{li2019deep} contains 2500 images of Drosophila embryos of 5 different categories, each of which contains 500 samples. One view is with 1750-dimensional visual features and the other view is with 79-dimensional textual features. \textbf{Caltech101-7}~\cite{dueck2007non} includes 5 visual feature descriptors, i.e., 40-dim wavelet moments~(WM) feature, 254-dim CENTRIST feature, 1,984-dim HOG feature, 512-dim GIST feature, and 928-dim LBP feature. These 1400 RGB images are sampled from 7 categories. Each category contains 200 images.
As for \textbf{COIL-20}, following Trosten \emph{et al.}~\shortcite{trosten2021reconsidering}, we establish a variant of COIL-20 with 480 object images divided into 20 classes, each of which is with 24 images. Every object is captured in 3 different poses with $128 \times 128$ pixel size for each view.
In terms of large-scale multi-view dataset, we apply \textbf{Multi-Fashion}~\cite{Xu_2022_CVPR}, which contains 10 types of clothes, e.g., pullover, shirt, and coat, to verify the scalability of the proposed CMHHC. In Multi-Fashion, different views of each instance are different fashionable designs of the same category, and there are 10000 $28 \times 28$ grey images in each view.

For all the above five datasets, we utilize the entire dataset of all samples and perform our model among all views. 
The statistics of the experimental datasets are summarized in Table~\ref{tab:dataset}.

\begin{table}[t]
\renewcommand{\arraystretch}{\tabelarraystretch}
    \centering
	\resizebox{0.95\columnwidth}{!}{
    \begin{tabular}{l|c|c|c|c|c}
        \toprule
        Dataset & MNIST-USPS & BDGP & Caltech101-7 & COIL-20 & Multi-Fashion\\
        \midrule
        \# Samples & 5000 & 2500 & 1400 & 480 & 10000 \\
        \# Categories & 10 & 5 & 7 & 20 & 10 \\
        \# views & 2 & 2 & 5 & 3 & 3 \\
        \bottomrule
    \end{tabular}
    }
    \caption{The statistics of the tested datasets.}
    \label{tab:dataset}
\end{table}

\subsection{Implementation Details}
\subsubsection{CMHHC}
Our entire model is implemented in the PyTorch platform. For representing the hierarchical structure efficiently, we use the corresponding SciPy, networkx, and ETE (Environment for Tree Exploration) Python toolkits. 
To speed up the convergence of our whole model, we first empirically pretrain $V$ autoencoders for 200 epochs on all datasets and the contrastive learning module for 10, 50, 50, 50 and 100 epochs on BDGP, MNIST-USPS, Caltech101-7, COIL-20 and Multi-Fashion respectively. Next, we finetune the whole multi-view representation learning process for 50 epochs. Finally, the epochs for HypHC training are set to 50. The batch size is set to $256$ and $512$ for multi-view representation learning and hierarchical clustering, using the Adam and hyperbolic-matched Riemannian optimizers~\cite{kochurov2020geoopt} respectively. The learning rate is set to~$5e^{-4}$ for Adam with~$256$ batch size, and~$5e^{-4}$ on BDGP, MNIST-USPS and Multi-Fashion,~$1e^{-3}$ on Caltech101-7 and COIL-20 for Riemannian Adam with~$512$ batch size. 
In addition,~$\tau$ is set to~$5e^{-1}$ for all datasets, while~$\tau_{c}$ is set to~$5e^{-2}$ for BDGP, MNIST-USPS and Multi-Fashion, and $1e^{-1}$ for Caltech101-7 and COIL-20.

In our CMHHC model, the network architecture consists of a  multi-view alignment learning module, a common-view similarity learning module, and a hyperbolic hierarchical clustering module.~(1) In multi-view alignment learning module, $V$ autoencoders are designed by the same architecture of full connected layers. Each encoder~$E^{(v)}(\cdot)$ is with dimensions of $D_\textrm{v}-500-500-2000-512$ and each decoder~$D^{(v)}(\cdot)$ is with dimensions of $512-2000-500-500-D_\textrm{v}$. The fully-connected contrastive learning layers~$f_{\textrm{con}}(\cdot)$ are with $512-512-218$ dimensions.~(2) In common-view similarity learning module, on concatenate aligned representations from all views, we adopt fully-connected similarity learning layers~$f_{\textrm{mom}}(\cdot)$ with $V\times D_\textrm{v}-128$ architecture.~(3) HypHC layers are implemented by~$f_{\textrm{hc}}(\cdot)$ with dimensions of $128-2$ to optimize hyperbolic embeddings\cite{chami2020trees}.
According to the general rule of hard mining strategy,~$k$ value is set to~$50$, and $k_{\textrm{pos}}$ value equals $N/2K$ and the~$k_{\textrm{neg}}$ value equals $N/K$ for all datasets. More specifically, ~$k_{\textrm{pos}}$ and~$k_{\textrm{neg}}$ values of 5 datasets are set as Table~\ref{tab:k}.

To achieve a tradeoff between time complexity and hierarchical clustering quality, the number of sampled triplets for HC input is empirically set to 900000, 3000000, 90000, 90000 and 9000000 for 5 datasets of different scales, i.e., BDGP, MNIST-USPS, Caltech101-7, COIL-20 and Multi-Fashion, respectively. The numbers of triplets sampled from datasets with more instances should be larger for expected clustering results~\cite{chami2020trees}.

\begin{table}[t]
\renewcommand{\arraystretch}{\tabelarraystretch}
    \centering
	\resizebox{0.95\columnwidth}{!}{
    \begin{tabular}{c|c|c|c|c|c}
        \toprule
        Dataset & MNIST-USPS & BDGP & Caltech101-7 & COIL-20 & Multi-Fashion \\
        \midrule
        $k_{\textrm{pos}}$ & 250 & 250 & 100 & 12 & 500 \\
        $k_{\textrm{neg}}$ & 500 & 500 & 200 & 24 & 1000 \\
        \bottomrule
    \end{tabular}
    }
    \caption{The~$k_{\textrm{pos}}$ and~$k_{\textrm{neg}}$ values of four datasets.}
    \label{tab:k}
\end{table}

\subsubsection{Baseline Methods}
For comparing conventional linkage-based discrete single-view HAC methods and continuous single-view HC methods fairly, we concatenate all views into a single view without losing information, and then apply the above methods. To be specific, we implemented HACs, i.e., Single-linkage, Complete-linkage, Average-linkage, and Ward-linkage algorithms by the corresponding SciPy Python library. We directly use the open-source implementations of UFit~\cite{chierchia2020ultrametric}
and HypHC~\cite{chami2020trees}. UFit proposed a series of continuous objectives for ultrametric fitting from different aspects. We follow the hyperparameter~$\lambda=10$ in UFit, and adopt the best cost function~``\emph{Closest+Size}'' instead of other proposed cost functions, i.e.,~``\emph{Closest+Triplet}'' and~``\emph{Dasgupta}''. The reasons for the choice of UFit cost function are twofold. First,~``\emph{Closest+Triplet}'' assumes the ground-truth labels of some data points are known to establish triplets for metric learning, which conflicts with the original intention of unsupervised clustering. Second, according to reported results by~\cite{chierchia2020ultrametric}, the performance of~``\emph{Dasgupta}'' is slightly worse than that of~`\emph{Closest+Size}'' in terms of accuracy
~(ACC). For HypHC, we adjust corresponding hyperparameters for 5 datasets of different scales. Here, for the sake of fairness, the number of sampled triplets for each dataset keeps consistent with that in our CMHHC model. Besides, we set $\tau_c=5e^{-1}$ for BDGP, MNIST-USPS and Multi-Fashion and $\tau_c=1e^{-1}$ for Caltech101-7 and COIL-20.

In terms of MHC~\cite{zheng2020multi}, we implemented it with Python as there was no open-source implementation available. MHC assumed that multiple views can be reconstructed by one fundamental latent representation~$\boldsymbol{H}$. However, a detailed explanation of latent representation is implicit. Hence, we empirically adopt non-negative matrix factorization on unified concatenate views to obtain~$\boldsymbol{H}$.
In addition, the way MHC built the adjacency graph for NNA was ambiguous, so we relax the formulation of the adjacency graph. We consider connecting three kinds of pairs~$(\boldsymbol{h}_i,\boldsymbol{h}_j)$ into one cluster, i.e., point~$\boldsymbol{h}_i$ is the nearest neighbor of point~$\boldsymbol{h}_j$, point~$\boldsymbol{h}_j$ is the nearest neighbor of point~$\boldsymbol{h}_i$ or points~$(\boldsymbol{h}_i,\boldsymbol{h}_j)$ have the same neighbor.

\subsection{Hierarchical Clustering Metrics}
Following~\cite{kobren2017hierarchical,monath2019gradient}, we validate hierarchical clustering performance via the Dendrogram Purity~(DP), which is a more holistic measurement of the hierarchical tree quality. 
Given a final clustering tree~$T$ of a dataset~$X$, and corresponding ground-truth clustering partitions~$\boldsymbol{K}=\left\{K_c\right\}_{c=1}^C$ belonging to~$C$ clusters, DP of~$T$ is defined as:
\begin{equation}
\small
{\rm DP}(T) =\frac{1}{|Ps|}\sum\limits_{c=1}^C\sum\limits_{(\boldsymbol{x}_i,\boldsymbol{x}_j)\in K_c\times K_c}{\rm pur}({\rm leaves}(T[i\vee j]),K_c)\,,
\end{equation}
where~$Ps=\left\{(\boldsymbol{x}_i,\boldsymbol{x}_j)|K(\boldsymbol{x}_i)=K(\boldsymbol{x}_j)\right\}$ represents two data points belonging to the same ground-truth cluster~$K_c$,~$\boldsymbol{x}_j$ and~${\rm leaves}(T[i\vee j])$ means the set of descendant leaves of LCA internal node~$i\vee j$, and~${\rm pur}(A,B)$ generally represents the proportion that the data points belonging to both set~$A$ and set~$B$ account for those belonging to set~$A$. Intuitively, high DP scores lead to nodes that are similar to clusters in the ground truth flat partition. 

\subsection{Hard Mining Strategy Analysis}
\begin{table}[t]
\renewcommand{\arraystretch}{\tabelarraystretch}
    \centering
	\resizebox{0.95\columnwidth}{!}{
    \begin{tabular}{c|c|c|c|c|c}
        \toprule
        Dataset & MNIST-USPS & BDGP & Caltech101-7 & COIL-20 & Multi-Fashion \\
        \midrule
        CMMC & \textbf{94.49\%} & \textbf{91.53\%} & \textbf{66.52\%} & \textbf{84.89\%} & \textbf{96.25\%} \\
        CMMC$_{\textrm{hard}}$ & 76.24\% & 83.96\% & 54.44\% & 82.05\% & 93.86\% \\
        \bottomrule
    \end{tabular}
    }
    \caption{Hard mining strategy analysis, i.e., DP results of CMMC and CMMC$_{\textrm{hard}}$.}
    \label{tab:hard}
\end{table}

\begin{figure*}[htbp]
	\centering
	\includegraphics[width=0.9\linewidth]{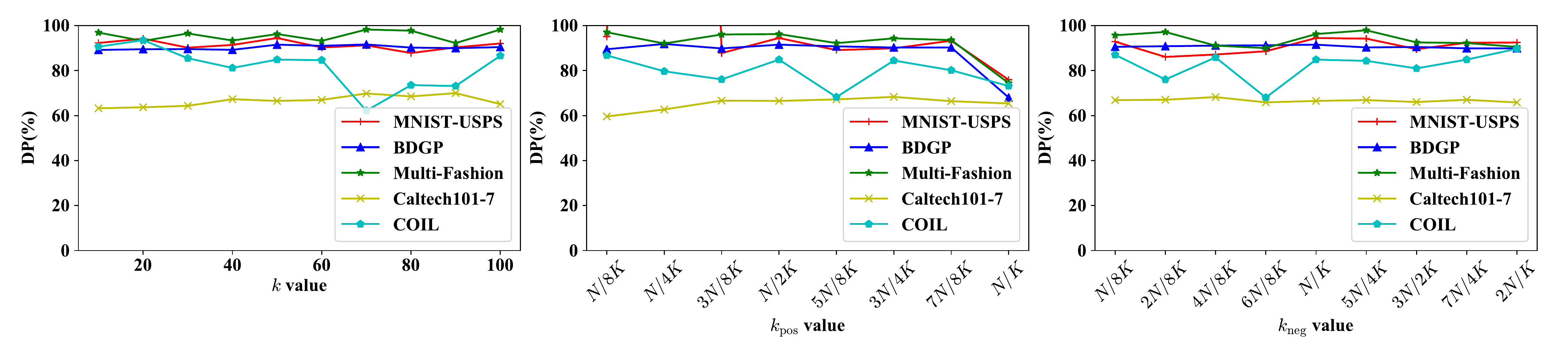}
	\caption{DP results of our CMHHC on five datasets with respect to different parameter settings.}
	\label{fig:parameter}
\end{figure*}

To demonstrate the effectiveness of the proposed hard mining strategy for unsupervised metric learning, we compare CMHHC with CMHHC$_{\textrm{hard}}$, denoting our model replacing proposed strategy with a much more strict triplet sampling strategy for weighted triplet loss training. In this way, the hard negative set for one anchor sample is made by involving nearest neighbors close to the anchor in the Euclidean space but on different manifold~\cite{iscen2018mining}. Therefore, we denote the hard negative set for~CMHHC$_{\textrm{hard}}$ as: 
\begin{equation}
\begin{aligned}
S_{\textrm{neg}}(\boldsymbol{h}_i) = \textrm{ENN}_{k_{\textrm{neg}}}(\boldsymbol{h}_i) - \textrm{MNN}_{k_{\textrm{neg}}}(\boldsymbol{h}_i)\,,
\end{aligned}
\end{equation}
The more strict triplet sampling strategy makes CMHHC$_{\textrm{hard}}$ focus on too hard negatives, which tend to be too close to the anchor sample in the Euclidean space. Therefore, mined pairwise similarity information is likely to contain unexpected noise, which will mislead the hierarchical clustering. Besides, CMHHC$_{\textrm{hard}}$ is limited to local structure around~$k_{\textrm{neg}}$ negatives, which does not respect to the fact that the whole clustering process takes the similarity of all over instances into consideration. However, our hard mining strategy, which regards samples not only far from the anchor on manifolds but also in the Euclidean space, is more applicable for our downstream clustering task. 

Table~\ref{tab:hard} shows the DP results of CMHHC and CMHHC$_{\textrm{hard}}$ from a quantitative perspective. Clearly, our hard mining strategy improves DP measurement on all datasets by a large margin. In other words, the relaxed strategy is capable of learning more meaningful triplets, which actually help generate clustering-friendly embedding space.

\subsection{Parameter Sensitivity Analysis}
The hyperparameters of CMHHC include the temperature parameters $\tau$, $\tau_\textrm{c}$ for contrastive learning and hyperbolic hierarchical clustering, and also, the number of nearest neighbors in the Euclidean space~$k$, and the values of hard positives and hard negatives $k_{\textrm{pos}}$ and $k_{\textrm{neg}}$ for similarity learning. Naturally, we set~$\tau=0.5$ for all datasets, while~$\tau_{\textrm c}=5e^{-2}$ for BDGP, MNIST-USPS, and Multi-Fashion and $\tau_\textrm{c}=1e^{-1}$ for Caltech101-7 and COIL-20, which are empirically efficient. In terms of the similarity learning parameters $k$, $k_{\textrm{pos}}$ and $k_{\textrm{neg}}$, we made the general rule that the~$k$ value equals $50$, $k_{\textrm{pos}}$ value equals $N/2K$ and the~$k_{\textrm{neg}}$ value equals $N/K$. Therefore, we evaluate the effectiveness of the general rule on all five datasets. With fixed the other two parameters~$\tau$ and~$\tau_c$, we vary $k$, $k_{\textrm{pos}}$ and $k_{\textrm{neg}}$ values in the range of $[10, \cdots, 100]$, $[N/8K, \cdots, N/K]$ and $[N/8K, \cdots, 2N/K]$ for all datasets. We also run the model 5 times, and the DP results with the lowest value of~$L_c$ under each parameter setting is shown in Fig~\ref{fig:parameter}. Our CMHHC is insensitive to $k$ value setting. Since the parameter $k$ is utilized to define the manifold similarity matrix, different $k$ values result in different metric properties on manifolds. Hence, the performance of CMHHC is likely to fluctuate within a reasonable range, and when $k=50$ the DP results on different datasets tend to be stable at a better level. 
Besides, setting $k_{\textrm{pos}}$ and $k_{\textrm{neg}}$ values via our general rule is easier to achieve fairly good hierarchical clustering performance. Actually, the parameters $k_{\textrm{pos}}$ and $k_{\textrm{neg}}$ controls the diversity and the difficulty of positives and negatives. More specifically, making the $k_{\textrm{neg}}$ equals $N/K$ and the $k_{\textrm{pos}}$ equals $N/2K$, offers sufficient hard positives and negatives, and guarantees these tuples to capture pseudo-label information with little noise.

\end{document}